\documentclass{article}

\usepackage{authblk}
\usepackage{times}
\usepackage[pdfborder={0 0 0}]{hyperref}

\title{Linear Bandit algorithms using the Bootstrap}

\author[1]{\textbf {Nandan Sudarsanam}}
\author[2]{\textbf {Balaraman Ravindran}}

\affil[1]{Department of Management Studies \\
	Indian Institute of Technology, Madras}
\affil[2]{Computer Science and Engineering \\
	Indian Institute of Technology, Madras}

\affil[ ]{\textbraceleft nandan\textbraceright @iitm.ac.in , \textbraceleft ravi\textbraceright@cse.iitm.ac.in}

\usepackage{ijcai16}
\usepackage{booktabs}%
\usepackage{algorithm}
\usepackage{algpseudocode}
\usepackage{graphicx}
\usepackage{float}
\usepackage{amsmath}
\usepackage{color,soul}

\begin{document}

\maketitle

\begin{abstract}
This study presents two new algorithms for solving linear stochastic bandit problems. The proposed methods use an approach from nonparametric statistics called bootstrapping to create confidence bounds. This is achieved without making any assumptions about the distribution of noise in the underlying system. We present the $X$-Random and $X$-Fixed bootstrap bandits which correspond to the two well-known approaches for conducting bootstraps on models, in the literature. The proposed methods are compared to other popular solutions for linear stochastic bandit problems, namely, OFUL, LinUCB and Thompson Sampling. The comparisons are carried out using a simulation study on a hierarchical probability meta-model, built from published data of experiments, which are run on real systems. The model representing the response surfaces is conceptualized as a Bayesian Network which is presented with varying degrees of noise for the simulations. One of the proposed methods, $X$-Random bootstrap, performs better than the baselines in-terms of cumulative regret across various degrees of noise and different number of trials. In certain settings the cumulative regret of this method is less than half of the best baseline. The $X$-Fixed bootstrap performs comparably in most situations and particularly well when the number of trials is low. The study concludes that these algorithms could be a preferred alternative for solving linear bandit problems, especially when the distribution of the noise in the system is unknown. 
\end{abstract}

\section{Introduction}

In its classical form, the multi-armed bandit (MAB) problem requires that a learning agent makes a choice, or selects an action, from $n$ alternatives, across $t$ trials, for each trial. After a choice is made in a given trial, the system presents the learning agent with a numerical reward from a stationary probability distribution associated with the action taken. The goal of the learning agent is to dynamically and sequentially choose alternatives, referred to as arms, which will maximize the expected total reward across the $t$ trials \cite{1}. An extended version of this problem statement is the linear bandit problem, which is sometimes referred through other names such as linear stochastic bandits or linear parameterized bandits, and sometimes studied under specific settings such as contextual or associative bandits with linear payoff functions. In its general form, originally proposed by Auer \shortcite{2} (whereas variants of the problem was considered prior to that \cite{22,23}), each arm or alternative is first parameterized into a set of features, which is known to the learning agent. The reward is still a sample from a stationary probability distribution, but its expected value is seen as the inner product of the feature vector and a fixed weight vector, which indicates the influence of each feature on the expected reward. The learning agent therefore seeks to understand these weights in its exploration phase, and uses this to exploit the system by picking promising arms. The critical advantage of this approach is that since we are tagging the rewards to the features, and not the arms directly, we do not need to pull each arm (the set of which could be large or even infinite), but would still be able to learn about the expected reward of each arm by understanding them through a common set of features (which is expected be smaller in size). \cite{2,3,4,5,6}.

The linear stochastic bandit conception allows us to extend various real-world online learning problems to a bandit framework. Studies by Abe and Long \shortcite{7} , Auer \shortcite{2}, and Li \textit{et al.}, \shortcite{6} discuss the application of a selection problem in displaying internet banner Ads or News articles. The agent needs to choose an Ad among many, to display to a specific user. Here the agent focuses its learning on understanding the features, which are combination of the User's and Ad's characteristics, as opposed to learning directly about each individual ad. The reward is received when a user clicks on the Ad. Similarly, Rusmevichientong and Tsitsikilis \shortcite{4} presents an application in marketing where the agent is tasked with choosing a product (arm) to offer to a customer, and the various product characteristics, such as price and popularity, are the features. The scope for application also goes beyond selection problems that span single objects or single processes conceptualized as arms (products, Ads). Situations exist where we have a set of decisions, or a combination of actions that need to be conceptually repackaged as a single arm. As discussed in Dani \textit{et al.}, \shortcite{3}, take the typical case of choosing one out of $K $ clinical treatments, which is modelled as a classical bandit problem with $ K$  arms. If we wanted to extend this context to a decision problem where we could choose any combination of the $K$ clinical treatments to be applied on a subject, then the set of possible decisions increase to $2^{K}$ (as there can be interactions between the treatments). In this situation, we could treat each of the $2^{K}$  combinations as separate arms in a linear bandit framework. We could conceptualize, at minimum, $ K$ features (corresponding to presence of absence of a treatment), and perhaps even some interaction terms and higher order transformations as additional features. Another novel application related to such combinatorics is discussed by Awerbuch and Kleinberg \shortcite{8} where an agent needs to figure out, online, the quickest path of getting from one point to another in a network. Here, multiple paths involving a combination of edges in the network can be selected to achieve this task and the agent learns the cost of choosing a certain path, but not the cost of each of the edges. Modelling this through linear bandit framework involves considering each path as an arm, and each edge as a feature. 

In this study, we propose two algorithms to solve the linear stochastic bandit problems without making any assumptions about the noise in the system. Inspired by practices in the statistics community to generate confidence bands on linear models through bootstraps \cite{10}, we apply these bootstrap algorithms in conjunction with our linear bandit formulation to create an arm pulling agent. Both these algorithms rely on the popular concept of using an upper confidence bound (UCB) associated with arms as described in Auer \shortcite{2}. The UCBs are derived from the bootstraps and used to choose an arm. The algorithms can also be conceptualized as an alternate way of building confidence sets as described in Abbasi-Yadkori \textit{et al.} \shortcite{5}. We compare these algorithms to other well researched solutions in the linear bandit space. Specifically, we look at the algorithm OFUL \cite{5}, LinUCB \cite{6}, and Thompson sampling for linear bandits \cite{9}. We perform this comparison using a simulation on a hierarchical probability meta-model, which create response surfaces inspired by real world systems.\\
 As described later in the Section 3, this approach allows us to choose the nature of the noise in the system. We experimented with different noise models, paying special attention to a case where the distributional assumptions of the state-of-the-art methods are violated. When we use Gaussian noise, which fit the assumptions made by the methods, the performance of our approach is comparable to the state-of-the-art. When we move away from this to something that is not sub-Gaussian (we chose the Laplace distribution in our study) the proposed methods start outperforming the state-of-the-art. 

The contributions of this study are as follows: 

1) This is the first ever use of bootstrap estimates of the confidence bounds in a bandit setup. Earlier contributions are limited to estimation of a posterior distribution or evaluation of solutions \cite{24,25}. The chief advantage of using bootstraps is that we make no distributional assumption about the stochasticity in the environment. We demonstrate the advantage of this by showing empirically that our approaches out perform state-of-the-art methods when the underlying assumptions are violated. 

2) While problems with combinatorial sets of decisions have been modelled as linear bandits earlier, the systems have been limited to modelling additive effects. We present a systematic method for including higher order interactions between choices. This allows us to expand the applicability of bandits to domains such as design of experiments where models with such higher order effects are necessary. 

3) The other contribution in this study pertains to the test environment used for the algorithms. Similar to other empirical studies on bandits, this study also uses synthetic data. However, the simulation data that is generated by our system is derived from real-world data gathered from published experiments on engineering systems. This meta-modelling approach allows us to capture idiosyncrasies of the real-world environments, but also perform simulations on a large number of response surfaces (as opposed to case studies). This study seeks to demonstrate the use of such an approach to testing bandit algorithms and also provide users with a specific response surface generating environment.

The paper is organized as follows. Section 2 introduces the proposed bootstrap algorithms, the notation and pseudo code, and the comparison algorithms. Section 3 describes the simulation environment. Section 4 presents the results with a discussion and identifies future research directions.

\section{Algorithms}
\subsection {Discussion}
\label{subsec:Discussion}

 Bootstrapping as a statistical approach seeks to sample from the data with replacement. 
This creates a version of the data which is different from the original set and therefore exploits the central idea of bootstrapping that \textit{The population is to the sample, as the sample is to the bootstrap sample’}. 
The two main algorithms presented in this study are the $X$-Fixed and $X$-random Bootstrap Bandits, corresponding to the similarly named approaches of bootstrapping to evaluate the confidence bands (different from confidence bounds as it applies to a model not a single parameter) of linear regression models discussed in the statistics literature. The use of these bootstrap approaches, especially with the above mentioned terminology, is first seen in Thompson \shortcite{10} and well discussed in various other sources \cite{11,12,13}.
 
 The algorithms presented in this study build on Auer \shortcite{2}'s idea of using the upper confidence bound (UCB) to create an arm pulling agent. The upper confidence bound is constructed using the bootstrap approach on linear regression as discussed above. The proposed algorithm can also be seen as an alternate way of implementing OFUL \cite{5}, which in turn also builds on the UCB idea of Auer\shortcite{2}. In trial $t$ the OFUL algorithm seeks to build the confidence set $C_{t-1}$ for the parameters or weight vector $\theta$, based off of rewards and arms pulled in previous rounds up to $t-1$. This set then serves as a constraint in the maximization of the inner product$ \langle x,\theta \rangle$, where $\theta \in C_{t-1}$ and $x \in U$, where $U$ is the set of all unique arms. The arm resulting from this maximization is then selected or pulled for round $t$. The confidence set is built by constructing an ellipsoid in the parameter space around the regularized least-squares estimate of $\theta$, such that with a high pre-defined probability $\theta$ lies in $C_{t-1}$. Our bootstrap approach does not create this ellipsoid but seeks to directly populate or create a confidence set by using the bootstraps to create multiple linear regression fits, and therefore multiple sets of parameters. This study adopts the same optimism-in-the-face-of-uncertainty optimisation to choose the arm $X_{t}$.

  In a bandit framework, the use of sampling directly from past data to create an arm pulling agent is rarely seen (Thompson sampling, for instance, samples from a distribution, which is created from the data). However, there are some notable exceptions \cite{17,24,25}. In Baranasi \textit{et al.}, \shortcite{17}  a form of sub-sampling without replacement is applied to the MAB problem. Whereas, the technical notes by Eckles and Kaptein \shortcite{24}, and Osband and Roy \shortcite{25} look at computationally efficient modifications to the core bootstrap idea and apply it to generate posterior distributions in Thompson sampling framework for the MAB problem. There are various advantages to using the bootstrap approaches. The main advantage being that these methods allow us to get confidence bounds without making any assumptions on the distributions or independence of variates, unlike some of the baseline methods. While such an approach could be computationally intensive, it could be very helpful in cases where the sampling distribution is unknown, or difficult to derive.

Similar to most bandit algorithms, the proposed solutions also require an initialization step. However, owing to the linear bandit conception, the algorithms described in this section can start by pulling a relatively small subset of the set of all unique arms. The number and selection of arms is based on the number of features the linear bandit algorithm seeks to estimate. At minimum the algorithm would need to pull more arms than the number of features. Also, the arms need to be carefully selected in order to ensure that the correlation between features (mulitcollinearity) is zero. This problem is well handled for discrete or discretized input spaces in offline design of experiments. A simple class of designs (sets of arms) that can satisfy the constraints discussed are \textit{Orthogonal Arrays}. These are matrices which, in addition to satisfying more restrictive mathematical properties, select sets of arms were each feature assumes each discrete state an equal number of times and the correlation between any two features is zero \cite{20}. In this study we propose the initialization step to be handled through orthogonal arrays of minimum size such that all parameters can be estimated.

\subsection{Notation and Steps}
\label{subsec:steps}

Assume a system where there are $M$ unique arms, and $F$ features that describe each of these arms. In this context we are interested in two different sets of arms, represented by two different sets of vectors. We define the first vector set as $U$ which is of size $M$ and the comprehensive set of all unique arms. Each vector is of length $F$ corresponding to the number of features. We can also define the vector set $X$, in matrix form, which represents the arms that have been tried, and for which rewards have been gathered. Matrix $X$ is of dimensions $t\times F$. Each row corresponds to the arm pulled at a particular trial or round, and after $t$ trials this matrix has $t$ rows and each arm is represented through its $F$ features. We present the vector of rewards as vector $R$ which is of also of size $t$ and corresponds to the rewards received from the each row of matrix $X$. In addition to this we use the parameter $\delta$ which is tuneable and describes the percentage level of the upper confidence bound, and therefore implicitly the degree of exploration versus exploitation.

In both algorithms the step of initialization by pulling the minimum, carefully selected arms to estimate the features is conducted in step 2. In algorithm 1, the key steps for bootstrapping are captured in step 5 through 7. In algorithm 2, the key steps for bootstrapping are captured in steps 7 and 8. The next two paragraphs discuss the algorithms in detail.

 In Algorithm 1, $X$-Random, for each trial we create a bootstrap sample of the data. Here, each input(arm)-output(reward) pair is selected randomly with replacement, creating a bootstrapped dataset of the same original size (steps 5-6). This dataset is regressed to estimate the parameters corresponding to that bootstrap (step 7). This creates a bootstrapped sample of parameters. For this sample we estimate the expected reward across all arms (loop in steps 8-10). We repeat these steps for various bootstraps of the residuals (loop in steps 4-11).The principle of upper confidence is then applied by tagging each arm with its "optimistic" performance (step 12). The arm with the best resulting reward is selected in that trial and receives a reward (step 14).  The updated input matrix and reward vector are used for subsequent trials (step 3). The algorithm ends after exhausting all trials (step 16).
 
  In the case of Algorithm 2, $X$-fixed, for each trial we fit a linear model to the historic data (step 4). The algorithm then computes a vector of residuals resulting from the difference between actual reward and the rewards expected from the linear model (step 5). The residuals are then bootstrapped to create new sample of residuals which are of the same size as the original set (step 7).Note that this is different from algorithm 1 which bootstraps the data points directly. We then apply the bootstrapped residuals to the fitted model to create new outputs, and re-regress the new outputs to the fixed inputs (step 8). The algorithm then proceeds identically to the $X$-Random algorithm.
 
 Literature suggests that the conceptualization of ‘what constitutes the sample’ should decide which of the two approaches we could use. If we believe that the inputs are fixed, and the output is a random sample from a distribution whose mean is determined by $X$, then the $X$-Fixed is appropriate. However, if we believe that the inputs characterized by the $X$ matrix is itself a random sample, then we would find the $X$-Random to be appropriate \cite{11}. Given that in our context the initialization step involves a carefully selected input matrix in the form of an orthogonal array, $X$-Fixed should be a better choice, initially. However, subsequent arm pulls are based off of the algorithms reacting to reward data which is conceptualized as a random sample taken from s stationary probability distribution and hence $X$-Random would seem to be an appropriate choice, asymptotically.

\subsection {Pseudo code}
\begin{algorithm}[H]
	\caption{$X$-Random Boostrap Bandit}\label{xrandomalgo}
	\begin{algorithmic}[1]
		
		\Procedure{XRNDBootstrapBandit}{$T$,$B$,$U$}
		\State Initialize with Orthogonal Array $X$ and obtain $R$
		\For {$t\leftarrow 1,T$}
		\For{$b\leftarrow 1,B$}
		\State Bootstrap rows of $X$ to create $X_b$ 
		\State Create $R_b$ from $R$ corresponding to $X_{b}$
		\State $\beta_{b}\leftarrow(X_{b}^{'} X_{b})^{-1} X_{b}^{'}(R_{b})$
		\For {$m\leftarrow 1,M$}
		\State $Y_{m,b}= U_(m)\times\beta_b$
		\EndFor
		\EndFor
		\State $Y_{\delta}(m)\leftarrow \delta^{th} percentile(Y_{m,.})$
		\State Select arm $U_{m}^{max}\leftarrow \max\limits_{m \in M} Y_{\delta}(m)$
		\State Receive reward $r_{t}$ from arm $U_{m}^{max}$ for trial $t$
		\State Append $X\leftarrow[X;U_{m}^{max}]$ and $R=[R;r_{t}]$
		\EndFor
		\EndProcedure
	\end{algorithmic}
\end{algorithm}

\begin{algorithm}
	\caption{$X$-Fixed Boostrap Bandit}\label{xfixedalgo}
	\begin{algorithmic}[1]
		
		\Procedure{XFixedBootstrapBandit}{$T$,$B$,$U$}
		\State Initialize with Orthogonal Array $X$ and obtain $R$
		\For {$t\leftarrow 1,T$}
		\State $\beta^{\ast}\leftarrow \lbrack X^{'}X\rbrack^{-1} X^{'} Y$
		\State $e^{\ast}\leftarrow R-X\beta^{\ast}$
		\For{$b\leftarrow 1,B$}
		\State Bootstrap $e^{\ast}$ to create $e_b$ 
		\State $\beta_{b}\leftarrow(X^{'} X)^{-1} X^{'}(X \beta^{\ast}+e_{b})$
		\For {$m\leftarrow 1,M$}
		\State $Y_{m,b}= U_(m)\times\beta_b$
	\EndFor
	\EndFor
	\State $Y_{\delta}(m)\leftarrow \delta^{th} percentile(Y_{m,.})$
	\State Select arm $U_{m}^{max}\leftarrow \max\limits_{m \in M} Y_{\delta}(m)$
	\State Receive reward $r_{t}$ from arm $U_{m}^{max}$ for trial $t$
    \State Append $X\leftarrow[X;U_{m}^{max}]$ and $R=[R;r_{t}]$
	\EndFor
	\EndProcedure
	\end{algorithmic}
\end{algorithm}

\subsection{Baseline Algorithms}

This study compares the bootstrapped linear bandits to three other established methods of working with linear bandits. These include OFUL \cite{5}, LinUCB \cite{6}, and Thompson sampling for linear bandits \cite{9}. The OFUL algorithm, owing to its extensibility to the proposed algorithms, has been discussed in detail in section~\ref{subsec:Discussion}. The LinUCB algorithm uses a similar conception as OFUL and other UCB based algorithms of using the uncertainty in the regularized least squares linear model that is built from past data to estimate the upper confidence bound. The Thompson sampling approach utilizes a Bayesian set up, where, in each step a parameter vector is sampled from posterior distributions of the parameters. The arm that maximizes the reward for this sampled vector is chosen, and a corresponding reward is received. The posterior distribution is then updated to account for the newly received reward.

\section{Testing Environment}
\label{sec:testbed}
Inspired by the examples discussed in \cite{3} and \cite{8}, we create a combinatorial application of linear bandits, and perform a simulation study on it. Specifically we replicate the extension of experimenting with clinical treatments. We take a sample case where $K=7$ treatments can be offered to patients, but these treatments are not mutually exclusive. Hence leading to $2^{k}=128$ possible combinations or arms to decide from. We parameterize a total of 28 features for the linear bandits, corresponding to the main effects of the $7$ treatments, and the $7C_{2}=21$  two-way interactions. Given this feature set, each bandit algorithm is seeded with an initial form of experimentation from a $2^{7-2}=32$ run factorial experiment, the minimum balanced design required to estimate all the $28$ features.

The response surfaces for experimentation are simulated based off of the general linear model, with main, effects, two-way interactions, three-way interactions, no higher order effects, and noise from the Laplace distribution, as shown in the equation below.\footnote{Similar experiments were conducted with Gaussian noise. As mentioned in the introduction, the results of the two bootstrap approaches were comparable to the state-of-the-art for different noise levels. Due to lack of space we are not reporting the results here.} 
\setlength{\abovedisplayskip}{0pt} \setlength{\abovedisplayshortskip}{0pt}
\begin{equation} \label{onlyeqn}
\begin{split}
 r_{t}= \beta_{0}+ \sum_{i=1}^{7}\beta_{i}x_{i} + \sum_{i=1}^{6}\sum_{j=i+1}^{7}\beta_{ij}x_{i}x_{j} + ... \\
\sum_{i=1}^{5}\sum_{j=i+1}^{6}\sum_{k=j+1}^{7}\beta_{ijk}x_{i}x_{j}x_{k}+ \epsilon_{t}\\
\textrm{where } \epsilon_{t} \sim f(x|0,b)=\frac{1}{2b} exp(-\frac{|x|}{b})
\end{split}
 \end{equation}
The selection of response surfaces which could contain three-way interactions was intentionally included despite the bandit parameterizations ending with two-way interactions. This was done to reflect the likely scenarios of missing features in any parameterization exercise. One could also think of these as unknown features.

 In this study we look at a simulation of $10,000$ response surfaces characterized by different versions of equation \ref{onlyeqn} that assume different values for the $\beta s$ and noise. In order to make these reward generation functions typical of the real-world systems that the bandits are likely to encounter, this study uses a hierarchical probability meta-model (HPM). The HPM, originally proposed in Chipman \textit{et al.}, \shortcite{18}, provides the mathematical structure to determine $\beta s$ . This structure seeks to capture certain mathematical regularities seen in real-world response surfaces. Specifically, three properties are sought after. They include, sparsity, hierarchy, and heredity. Sparsity indicates that only a subset of the potential set of features or inputs will be found to have a statistically significant effect on the response. Hierarchy indicates that main effects tend to have larger co-efficients than two-way interactions, and two-way interactions tend to be larger than three-way, and so on. Heredity refers to the fact that the likelihood of an interaction having a significant effect on the response is affected by whether it's parents were significant, or not.
 
\begin{figure}\label{BN}
\centering
\includegraphics[width=0.65\linewidth]{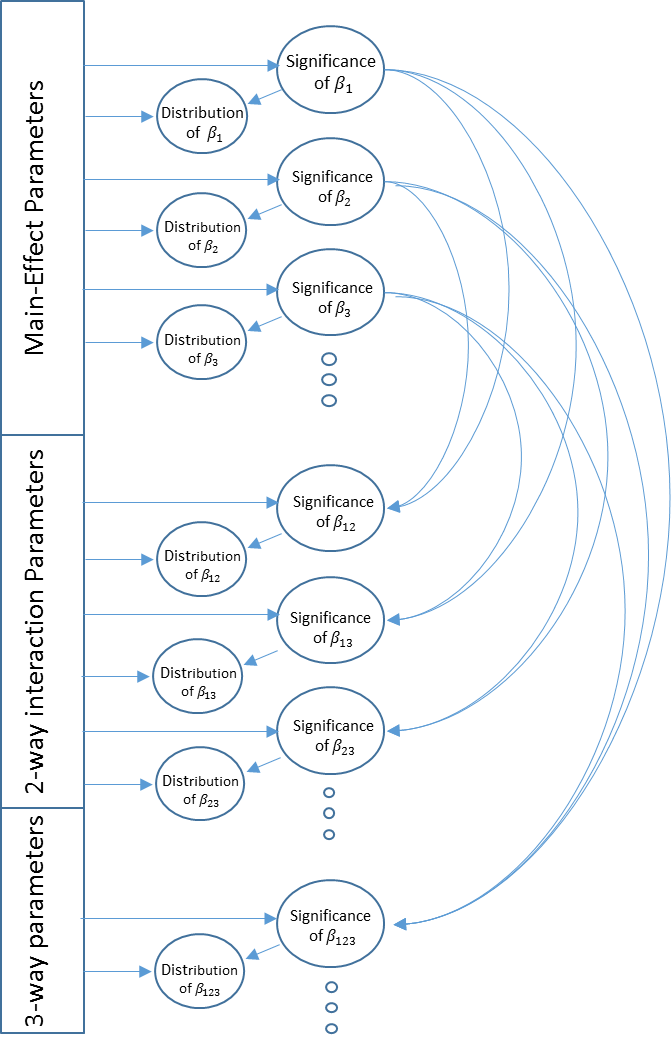}
\caption{Bayesian Network Representation of the HPM}
\label{fig:screenshot004}
\end{figure}

 Capturing these properties in the HPM requires modelling various parameters as random variables with conditional dependencies. This can represented as a Bayesian Network (BN) \cite{21}. The structure of using the BN to implement the HPM is illustrated through figure \ref{fig:screenshot004}. This representation shows the probabilistic relationships in a system with $3$ main-effects and their resulting interactions. However, in our simulation we use a system with $7$ main effects. The BN helps us determine the statistical significance of the various $\beta$s, which in turn help us implement the principle of sparsity and heredity. The BN also helps us infer the distributions of the $\beta$s which helps us implement hierarchy (the co-efficient is modelled as a random variable).

 \setlength{\textfloatsep}{10pt plus 1.0pt minus 2.0pt}
 A meta study by Li \textit{et al.}, \shortcite{14} adapts this HPM structure in conjunction with real-world data to estimate the parameters shown in Figure \ref{fig:screenshot004} . The parameters for the distribution and significance of the co-efficients is based off of $113$ data sets gathered from published academic work of engineering systems, spanning different disciplines. The adapted mathematical structure of HPMs can be seen in Frey and Li \shortcite{15}. The use of such a test bed to evaluate experimental algorithms can also been seen in other studies \cite{15,16}. The use of real-world meta data inspires our study to an environment of only $128$ arms for the linear bandits problem, as opposed to larger set of arms seen in other studies which use purely synthetic data.

\section{Results and Discussion}
\setlength{\belowdisplayskip}{0pt} \setlength{\belowdisplayshortskip}{0pt}
\setlength{\abovedisplayskip}{0pt} \setlength{\abovedisplayshortskip}{0pt}
The pseudo-performance of the selected arm, in line with pseudo-regret discussed in \cite{19}), is described by:
\begin{equation} \label{pseudoperf}
\frac{(U(m)_{t}.\theta)}{(U(m^{*}).\theta)}.100
\end{equation}
where $U(m)_{t}$ is the arm selected at trial $t$ which is represented as a vector of its feature values, $U(m^{*})$ is the optimal arm. The true parameters or feature weights are represented by $\theta$ (which are actually unknown to the agents) and described by their extended feature space of the intercept, the $7$ main effects, $21$ two-way interactions and $35$ three-way interactions ($64$ in total), since they represent the true underlying system. The algorithms (or agent), however, only tries to estimate the partial set of parameters discussed in section \ref{sec:testbed}.

We report our findings across three levels of noise, low ($\sigma_{\epsilon}=1$), medium($\sigma_{\epsilon}=5$), and high ($\sigma_{\epsilon}=10$), where the noise, $\epsilon$ is distributed as a Laplace random variable as shown in equation \ref{onlyeqn}.\footnote {As a reference for why we consider these levels as low, medium and high, the distribution of the significant main effects after scaling of the meta data is $N(0,10^{2})$ and the probability of the main effect being significant is $.41$. 
Hence at the high levels, the strength of the noise is as high as the the significant main-effects, and similar ratios can be derived for medium and low noise levels.}
	\setlength{\intextsep}{5pt plus 1.0pt minus 2.0pt}
\setlength{\abovecaptionskip}{1pt plus 1pt minus 2pt}

\begin{figure}[ht]
\centering
\includegraphics[width=1\linewidth]{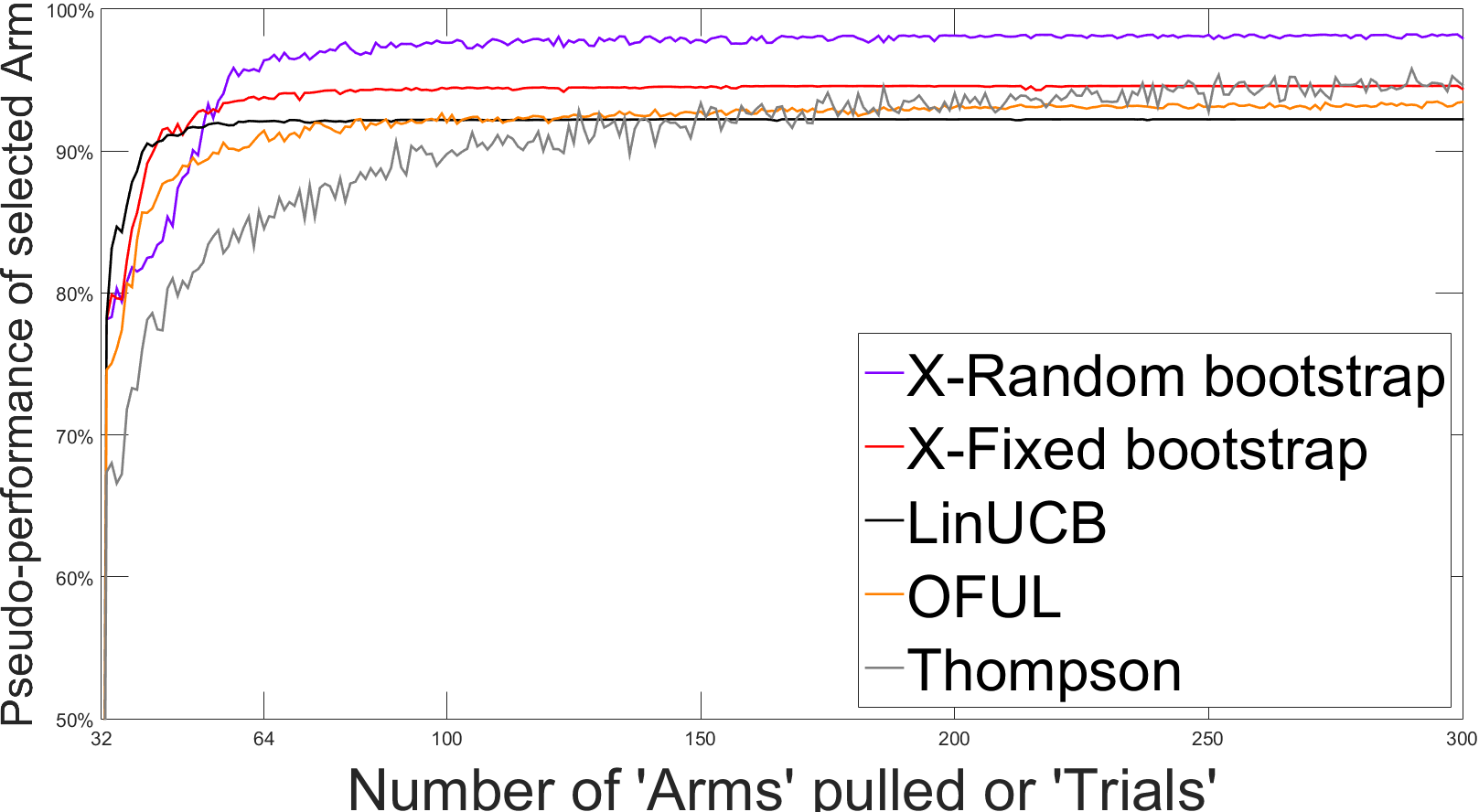}
\caption{Results across $10,000$ response surfaces for  $\sigma_{\epsilon}=1$}
\label{fig:1}
\end{figure}
\begin{figure}[H]
\centering
\includegraphics[width=1\linewidth]{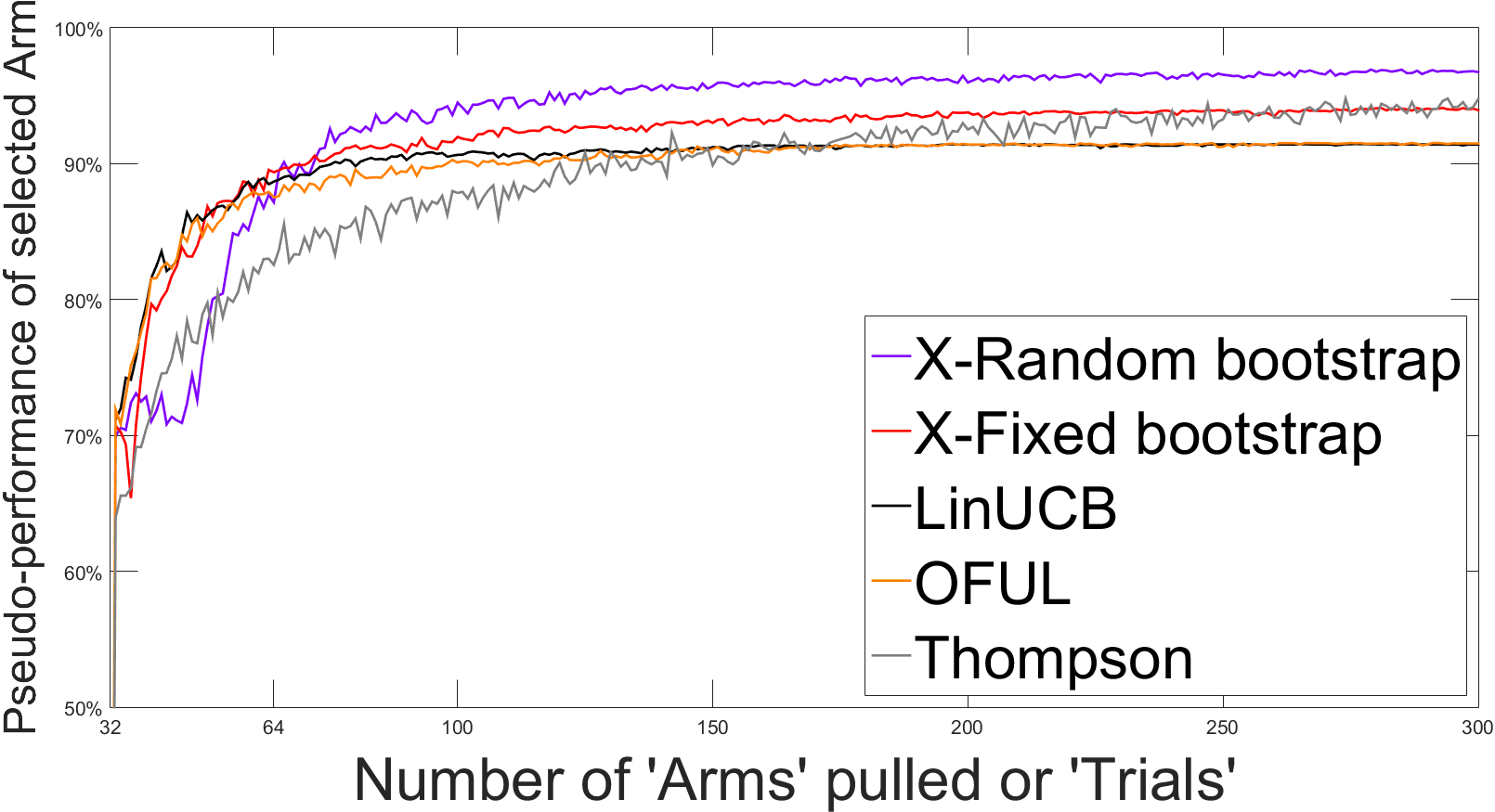}
\caption{Results across $10,000$ response surfaces for  $\sigma_{\epsilon}=5$}
\label{fig:2}
\end{figure}
\begin{figure}[H]
\centering
\includegraphics[width=1\linewidth]{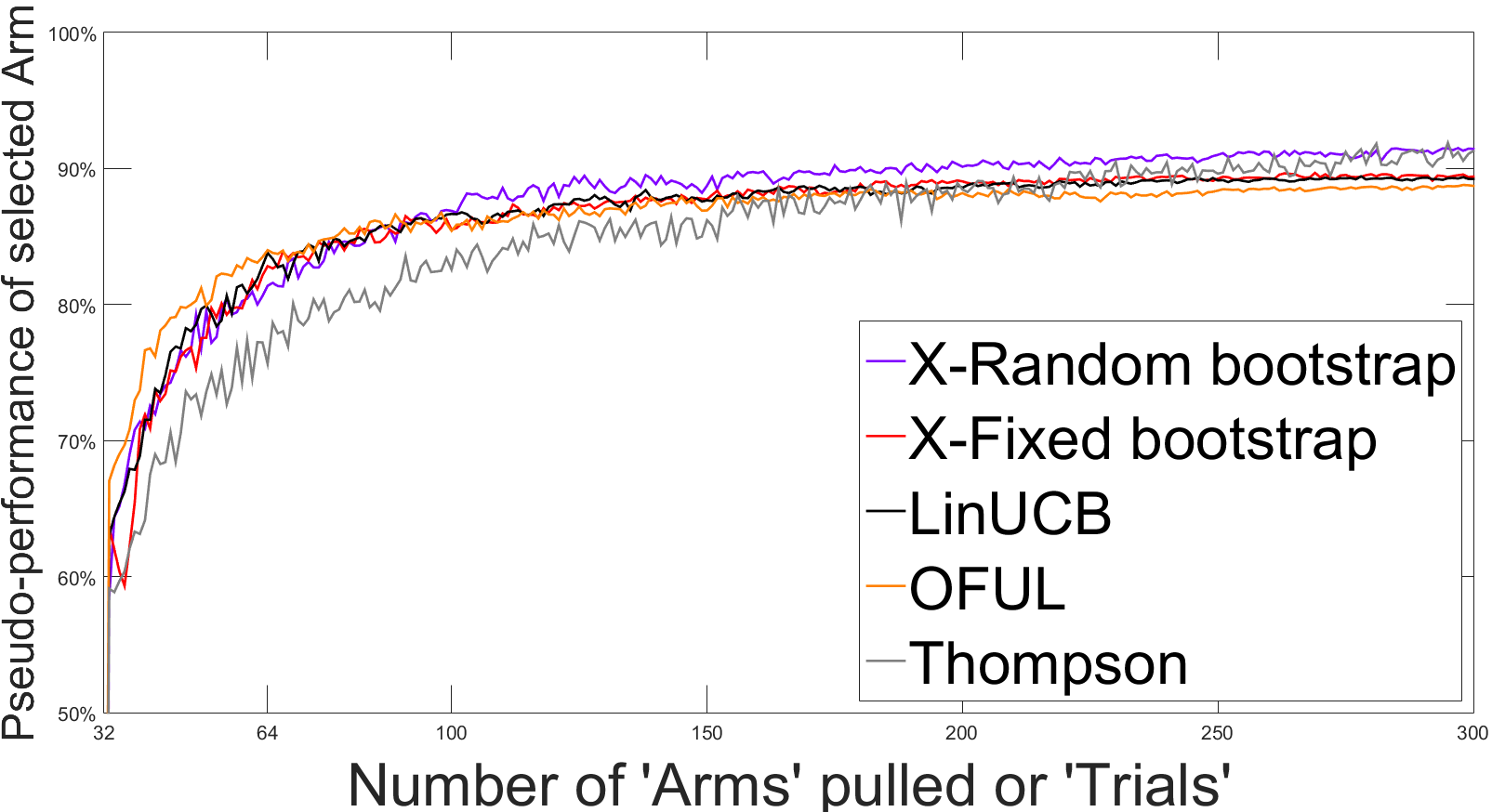}
\caption{Results across $10,000$ response surface for  $\sigma_{\epsilon}=10$}
\label{fig:3}
\end{figure}
  The implementation of X-Random, X-Fixed, OFUL and LinUCB require making some guesses on tuneable parameters which will influence the performance of the algorithm. The insight on what values are ideal need not be available to the agent, \textit{a}priori. However, it might be possible to make educated guesses based on the time horizon (number of trials), and degree of noise. To represent these algorithms and their baselines in the best case scenario, their parameters were fine tuned in the simulations. This was achieved with the objective of minimizing cumulative pseudo-regret over the 300 trial runs shown in the Figures \ref{fig:1}, \ref{fig:2}, and \ref{fig:3}. In this study, we also wanted to look at the effect of different number of trials, its direct impact on the algorithms as well as its interactive effects with the various noise levels. Table \ref{tbl:results} shows this. The cumulative regret shown in the table is an extension of equation \ref{pseudoperf} defined by $\sum_{t=1}^{T}1-\frac{(U(m)_{t}.\theta)}{(U(m^{*}).\theta)}$

 \begin{table}[H]
 	\centering
 	\includegraphics[width=1\linewidth]{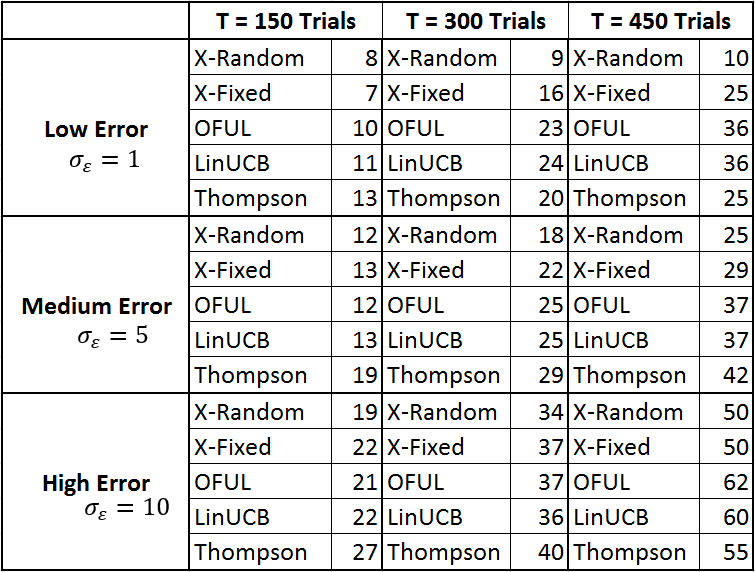}
 	\caption{Cumulative Regret for multiple time horizons}
 	\label{tbl:results}
 \end{table}
 
The results show that across all tested degrees of noise and all time horizons, one of the proposed methods, the $X$-Random bootstrap, shows lower cumulative regret than all the baseline methods. In certain environments, such as medium to high levels of noise, and lower number of trials, the difference between $X$-Random and the comparisons is marginal.  However, in other environments, such as low to medium levels of noise, and a larger number of trials, the difference is substantial. The cumulative regret in such cases for the $X$-Random is less than $1/2$ of the best baseline approach. In general, the $X$-Fixed bootstrap also outperforms the baselines but this is not consistent across all the tested dimensions. Similar to the competitive advantage trend between $X$-Random and the baselines, the $X$-Fixed also performs relatively better and in lower noise settings and larger number of trials.

 In terms of the relative performances between the bootstrap approaches, we see that in general the $X$-Random outperforms $X$-fixed across the settings that were tested. However, it is noteworthy that $X$-Fixed comes closest to the $X$-Random for lower trial runs and even outperforms the $X$-Random at the lowest noise level for the lower trial setting.  As discussed in section \ref{subsec:steps}, this is to be expected and is due to the fact that the initial distribution of X is determined from a designed experiment (which is in line with the assumptions of $X$-Fixed) and not a random sample. However, subsequent arm selections are the result of inferences from the reward, which is a random draw from a distribution. This favours the X-Random in subsequent trials.
 

In addition to the comparison of the proposed algorithms with the baselines, the unique test-bed proposed in this study could also throw some light on the relative performances of the baseline algorithms when the distributional assumptions used by them are violated. Thompson sampling, when compared to OFUL and LinUCB, consistently performs poorly when the number of trials is small and performs best for longer time-horizons. The poor performance for the shorter trial limits is due to the fixed level of exploration in this algorithm which cannot be fine-tuned. However, for longer time horizons Thompson sampling outperforms OFUL and LinUCB even after the latter two have had their parameters fine-tuned. For the greater part, both LinUCB and OFUL perform similarly. This should also be expected owing to the similarities in the algorithms, the distributional assumptions, and the nature of the tuning parameters. However, small differences were seen between the two algorithms, with OFUL working better with fewer trials and lower noise, where as LinUCB did better with higher noise and a longer trial horizon.


 The study concludes that when the distribution of noise is not known, and there are no computational constraints, then the $X$-Random and $X$-fixed bootstraps can be a preferred alternative to currently popular linear bandit algorithms.

\subsection{Future Work}
Broadly, our future efforts seek to study the role of using distribution-free algorithms to increase the performance and applicability of bandit formulations. To that end, we would like to look at the use of non-parametric statistics techniques to solve various bandit extensions. We would also like to understand the specific situations or environments where the popular methods which make distributional assumptions tend to fail or underperform. 

Specifically, our next steps would be to create analytical regret bounds for the linear bootstrap approaches and look at efficient implementations of making the bootstrapped algorithms faster. 



\appendix

\bibliographystyle{named}
\bibliography{ijcai16}

\end{document}